\documentclass{article} 
\usepackage{iclr2016_conference}
\usepackage{times}
\usepackage{hyperref}
\usepackage{url}
\usepackage{amsmath}
\usepackage{amssymb}
\usepackage{amsfonts}
\usepackage{graphicx}
\usepackage{wrapfig}
\usepackage{algorithm}
\usepackage{algorithmicx}
\usepackage{algpseudocode}
\algtext*{EndWhile}
\algtext*{EndIf}
\algtext*{EndFunction}
\usepackage{caption}
\usepackage{subcaption}

\title{Neural Programmer-Interpreters}

\author{
Scott Reed $\&$ Nando de Freitas  \\
Google DeepMind \\
London, UK \\
\texttt{scott.ellison.reed@gmail.com} \\
\texttt{nandodefreitas@google.com} \\
}

\iclrfinalcopy 
\begin{document}
\maketitle
\begin{abstract}
We propose the neural programmer-interpreter (NPI): a recurrent and compositional neural network that learns to represent and execute programs.
NPI has three learnable components: a task-agnostic recurrent core, a persistent key-value program memory, and domain-specific encoders that enable a single NPI to operate in multiple perceptually diverse environments with distinct affordances.
%
%
By learning to compose lower-level programs to express higher-level programs, NPI reduces sample complexity and increases generalization ability compared to sequence-to-sequence LSTMs.
%
The program memory allows efficient learning of additional tasks by building on existing programs.
NPI can also harness the environment (e.g. a scratch pad with read-write pointers) to cache intermediate results of computation, lessening the long-term memory burden on recurrent hidden units.
In this work we train the NPI with fully-supervised execution traces; each program has example sequences of calls to the immediate subprograms conditioned on the input.
%
Rather than training on a huge number of relatively weak labels, NPI learns from a small number of rich examples.
We demonstrate the capability of our model to learn several types of compositional programs: addition, sorting, and canonicalizing 3D models. 
Furthermore, a \emph{single} NPI learns to execute these programs and all 21 associated subprograms.

\end{abstract}
\vspace{-0.1in}
\section{Introduction}
\vspace{-0.1in}
Teaching machines to learn new programs, to rapidly compose new programs from existing programs, and to conditionally execute these programs automatically so as to solve a wide variety of tasks is one of the central challenges of AI. Programs appear in many guises in various AI problems; including motor behaviours, image  transformations, reinforcement learning policies, classical algorithms, and symbolic relations.

\begin{figure}[t!]
  \includegraphics[width=\textwidth]{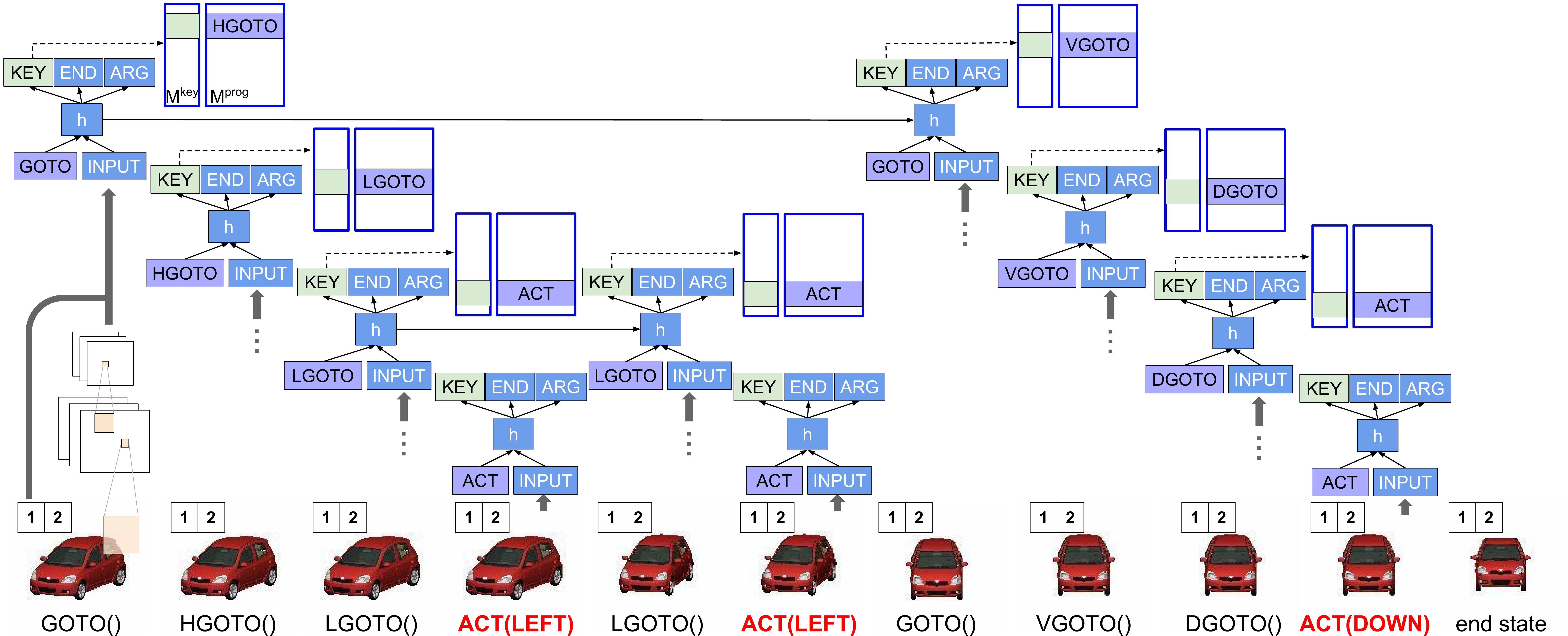}
  \caption{Example execution of canonicalizing 3D car models. The task is to move the camera such that a target angle and elevation are reached. There is a read-only scratch pad containing the target (angle 1, elevation 2 here). The image encoder is a convnet trained from scratch on pixels.}
  \vspace{2mm}
  \begin{minipage}[c]{0.67\textwidth}
    \includegraphics[width=\textwidth]{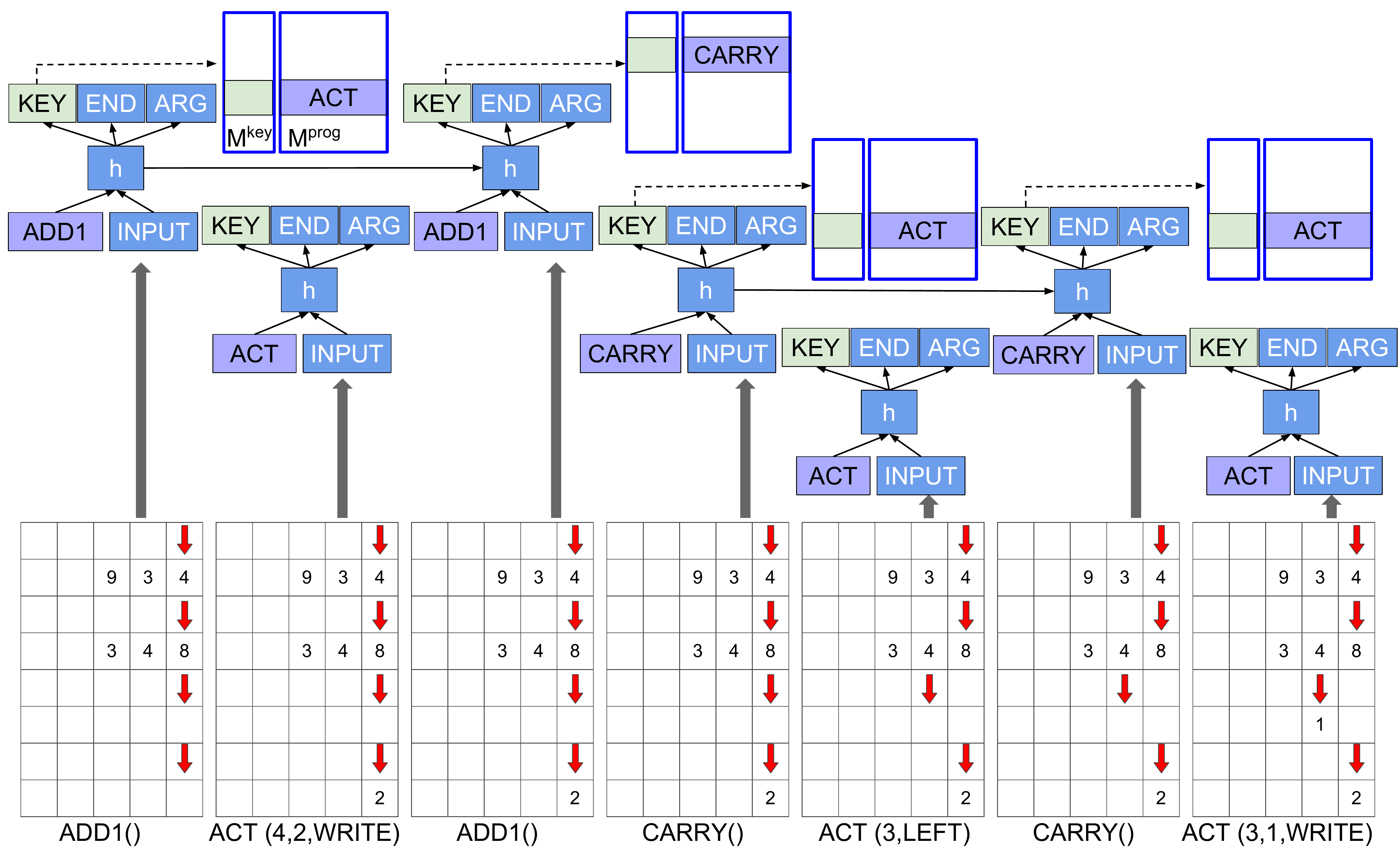}
  \end{minipage}\hfill
  \begin{minipage}[c]{0.3\textwidth}
    \vspace{0.1in}
    \caption{Example execution trace of single-digit addition. The task is to perform a single-digit add on the numbers at pointer locations in the first two rows. The carry (row 3) and output (row 4) should be updated to reflect the addition.
At each time step, an observation of the environment (viewed from each pointer on a scratch pad) is encoded into a fixed-length vector.
%
%
}
\label{fig:03-03}
  \end{minipage}
  \label{fig:model}
  \vspace{0.0in}
\end{figure}

In this paper, we develop a compositional architecture that learns to represent and interpret programs.
We refer to this architecture as the Neural Programmer-Interpreter (NPI).
The core module is an LSTM-based sequence model that takes as input a learnable program embedding, program arguments passed on by the calling program, and a feature representation of the environment. The output of the core module is a key indicating what program to call next, arguments for the following program and a flag indicating whether the program should terminate.
In addition to the recurrent core, the NPI architecture includes a learnable key-value memory of program embeddings.
This program-memory is essential for learning and re-using programs in a continual manner. Figures~\ref{fig:model} and~\ref{fig:03-03} illustrate the NPI on two different tasks. 

We show in our experiments that the NPI architecture can learn 21 programs, including addition, sorting, and trajectory planning from image pixels. Crucially, this can be achieved using a single core model with the same parameters shared across all tasks.
Different environments (for example images, text, and scratch-pads) may require specific perception modules or encoders to produce the features used by the shared core, as well as environment-specific actuators. Both perception modules and actuators can be learned from data when training the NPI architecture.
%

%
To train the NPI we use curriculum learning and supervision via example execution traces.
Each program has example sequences of calls to the immediate subprograms conditioned on the input. 
By using neural networks to represent the subprograms and learning these from data, the approach can generalize on tasks involving rich perceptual inputs and uncertainty.

We may envision two approaches to provide supervision. In one, we provide a very large number of labeled examples, as in object recognition, speech and machine translation. In the other, the approached followed in this paper, the aim is to provide far fewer labeled examples, but where the labels contain richer information allowing the model to learn compositional structure. While unsupervised and reinforcement learning play important roles in perception and motor control, other cognitive abilities are possible thanks to rich supervision and curriculum learning. This is indeed the reason for sending our children to school.


An advantage of our approach to model building and training is that the learned programs exhibit \emph{strong generalization}. Specifically, when trained to sort sequences of up to twenty numbers in length, they can sort much longer sequences at test time. In contrast, the experiments will show that more standard sequence to sequence LSTMs only exhibit \emph{weak generalization}, see Figure~\ref{fig:sc_gen}.

A trained NPI with fixed parameters and a learned library of programs, can act both as an interpreter and as a programmer. As an interpreter, it takes input in the form of a program embedding and input data and subsequently executes the program. As a programmer, it uses samples drawn from a new task to generate a new program embedding that can be added to its library of programs.
\vspace{-0.1in}
\section{Related work}
\vspace{-0.1in}
%
Several ideas related to our approach have a long history. 
For example, the idea of using dynamically programmable networks in which the activations of one network become the weights (the program) of a second network was mentioned in the Sigma-Pi units section of the influential PDP paper \citep{Rumelhart:1986}.
This idea appeared in \citep{Sutskever:2008} in the context of learning higher order symbolic relations and in \citep{Donnarumma:2015} as the key ingredient of an architecture for prefrontal cognitive control.
\citet{Schmidhuber:1992} proposed a related meta-learning idea, whereby one learns the parameters of a slowly changing network, which in turn generates context dependent weight changes for a second rapidly changing network. These approaches have only been demonstrated in very limited settings.
In cognitive science, several theories of brain areas controlling other brain parts so as to carry out multiple tasks have been proposed; see for example \citet{Schneider:2003,Anderson:2010} and \citet{Donnarumma:2012}.

Related problems have been studied in the literature on hierarchical reinforcement learning (\emph{e.g.}, ~\citet{Dietterich:2000,Andre:2000,Sutton:1999} and~\citet{Schaul:2015}), imitation and apprenticeship learning (\emph{e.g.}, \citet{Kolter:2008} and~\citet{Rothkopf:2013}) and elicitation of options through human interaction \citep{Subramanian:2011}.
These ideas have held great promise, but have not enjoyed significant impact.
We believe the recurrent compositional neural representations proposed in this paper could help these approaches in the future, and in particular in overcoming feature engineering.

Several recent advancements have extended recurrent networks to solve problems beyond simple sequence prediction.
%
\citet{graves2014neural} developed a neural Turing machine capable of learning and executing simple programs such as repeat copying, simple priority sorting and associative recall.
%
%
%
~\cite{vinyals2015pointer} developed Pointer Networks that generalize the notion of encoder attention in order to provide the decoder a variable-sized output space depending on the input sequence length.
This model was shown to be effective for combinatorial optimization problems such as the traveling salesman and Delaunay triangulation.
%
%
While our proposed model is trained on execution traces instead of input and output pairs, in exchange for this richer supervision we benefit from compositional program structure, improving data efficiency on several problems.

%
%

This work is also closely related to program induction.
Most previous work on program induction, i.e. inducing a program given example input and output pairs, has used genetic programming~\citep{banzhaf1998genetic} to evolve useful programs from candidate populations.
%
%
\citet{Mou:2014} process program symbols to learn max-margin program embeddings with the help of parse trees.
\cite{zaremba2014learning} trained LSTM models to read in the text of simple programs character-by-character and correctly predict the program output.
~\citet{joulin2015inferring} augmented a recurrent network with a pushdown stack, allowing for generalization to longer input sequences than seen during training for several algorithmic patterns.

Contemporary to this work, several papers have also studied program induction with variants of recurrent neural networks~\citep{zaremba2015reinforcement,zaremba2015learning,kaiser2015neural,kurach2015neural,neelakantan2015neural}.
While we share a similar motivation, our approach is distinct in that we explicitly incorporate compositional structure into the network using a program memory, allowing the model to learn new programs by combining sub-programs.
\vspace{-0.1in}
\section{Model}
\vspace{-0.1in}
%
The NPI core is a long short-term memory (LSTM) network \citep{hochreiter1997long} that acts as a router between programs conditioned on the current state observation and previous hidden unit states.
At each time step, the core module can select another program to invoke using content-based addressing.
It emits the probability of ending the current program with a single binary unit.
If this probability is over threshold (we used 0.5), control is returned to the caller by popping the caller's LSTM hidden units and program embedding off of a program call stack and resuming execution in this context.

The NPI may also optionally write arguments (ARG) that are passed by reference or value to the invoked sub-programs.  
For example, an argument could indicate a specific location in the input sequence (by reference), or it could specify a number to write down at a particular location in the sequence (by value).
The subsequent state consists of these arguments and observations of the environment. The approach is illustrated in Figures~\ref{fig:model} and~\ref{fig:03-03}. 
%
%
%
%
%
%
%
%
%

It must be emphasized that there is a single inference core.
That is, all the LSTM instantiations executing arbitrary programs share the same parameters.
Different programs correspond to program embeddings, which are stored in a learnable persistent memory.
%
%
The programs therefore have a more succinct representation than neural programs encoded as the full set of weights in a neural network \citep{Rumelhart:1986,graves2014neural}.

The output of an NPI, conditioned on an input state and a program to run, is a sequence of actions in a given environment.
In this work, we consider several environments: a 1-D array with read-only pointers and a swap action, a 2-D scratch pad with read-write pointers, and a CAD renderer with controllable elevation and azimuth movements. 
Note that the sequence of actions for a program is not fixed, but dependent also on the input state.
\vspace{-0.1in}
\subsection{Inference}
\vspace{-0.05in}
%
Denote the environment observation at time $t$ as $e_{t} \in \mathcal{E}$, and the current program arguments as $a_{t} \in \mathcal{A}$.
The form of $e_{t}$ can vary dramatically by environment; for example it could be a color image or an array of numbers.
%
The program arguments $a_{t}$ can also vary by environment, but in the experiments for this paper we always used a 3-tuple of integers $(a_t(1), a_t(2), a_t(3))$.
Given the environment and arguments at time $t$, a fixed-length state encoding $s_t \in \mathbb{R}^{D}$ is extracted by a domain-specific encoder $f_{enc} : \mathcal{E} \times \mathcal{A} \rightarrow \mathbb{R}^{D}$.
In section~\ref{sec:experiments} we provide examples of several encoders.
Note that a single NPI network can have multiple encoders for multiple environments, and encoders can potentially also be shared across tasks. 

We denote the current program embedding as $p_t \in \mathbb{R}^{P}$.
The previous hidden unit and cell states are $h^{(l)}_{t-1} \in \mathbb{R}^{M}$ and $c^{(l)}_{t-1} \in \mathbb{R}^{M}$, $l = 1,...,L$ where $L$ is the number of layers in the LSTM.
%
%
%
The program and state vectors are then propagated forward through an LSTM mapping $f_{lstm}$ as in \citep{sutskever2014sequence}. 
How to fuse $p_{t}$ and $s_{t}$ within $f_{lstm}$ is an implementation detail, but in this work we concatenate and feed through a 2-layer MLP with rectified linear (ReLU) hidden activation and linear decoder.

From the top LSTM hidden state $h_{t}^{L}$, several decoders generate the outputs.
The probability of finishing the program and returning to the caller~\footnote{In our implementation, a program may first call a subprogram before itself finishing. The only exception is the ACT program that signals a low-level action to the environment, e.g. moving a pointer one step left or writing a value. By convention ACT does not call any further sub-programs.} is computed by $f_{end} : \mathbb{R}^{M} \rightarrow [0,1]$.
The lookup key embedding used for retrieving the next program from memory is computed by $f_{prog} : \mathbb{R}^{M} \rightarrow \mathbb{R}^{K}$.
Note that $\mathbb{R}^{K}$ can be much smaller than $\mathbb{R}^{P}$ because the key only need act as the identifier of a program, while the program embedding must have enough capacity to conditionally generate a sequence of actions.
The contents of the arguments to the next program to be called are generated by $f_{arg} : \mathbb{R}^{M} \rightarrow \mathcal{A}$.
%
The feed-forward steps of program inference are summarized below:
\begin{align}
s_{t} &= f_{enc}(e_{t}, a_{t}) \\
h_{t} &= f_{lstm}(s_{t}, p_{t}, h_{t-1}) \label{eq:lstm} \\
r_{t} &= f_{end}(h_{t}), \text{  } k_{t} = f_{prog}(h_{t}), \text{  } a_{t+1} = f_{arg}(h_{t}) 
\end{align}
where $r_t$, $k_t$ and $a_{t+1}$ correspond to the end-of-program probability, program key embedding, and output arguments at time $t$, respectively.
These yield input arguments at time $t+1$. To simplify the notation, we have abstracted properties such as layers and cell memory in the sequence-to-sequence LSTM of equation~(\ref{eq:lstm}); see \citep{sutskever2014sequence} for details.

The NPI representation is equipped with key-value memory structures $M^{\texttt{key}} \in \mathbb{R}^{N \times K}$ and $M^{\texttt{prog}} \in \mathbb{R}^{N \times P}$ storing program keys and program embeddings, respectively, where $N$ is the current number of programs in memory.
We can add more programs by adding rows to memory. 

During training, the next program identifier is provided to the model as ground-truth, so that its embedding can be retrieved from the corresponding row of $M^{\texttt{prog}}$.
At test time, we compute the ``program ID'' by comparing the key embedding $k_{t}$ to each row of $M^{\texttt{key}}$ storing all program keys.
Then the program embedding is retrieved from $M^{\texttt{prog}}$ as follows:
%
\begin{align}
&i^{*} = \underset{i = 1 .. N}{\text{arg max}} (M^{\texttt{key}}_{i,:})^{T} k_{t} \texttt{   , }
p_{t+1} = M^{\texttt{prog}}_{i^{*},:}
\end{align}
The next environmental state $e_{t+1}$ will be determined by the dynamics of the environment and can be affected by both the choice of program $p_{t}$ and the contents of the output arguments $a_{t}$, \emph{i.e.}
\begin{align}
e_{t+1} \sim f_{env}(e_{t}, p_{t}, a_{t})
\end{align}
The transition mapping $f_{env}$ is domain-specific and will be discussed in Section~\ref{sec:experiments}.
A description of the inference procedure is given in Algorithm~\ref{alg:inference}. 

  \begin{algorithm}
    \caption{Neural programming inference\label{alg:inference}}
    \begin{algorithmic}[1]
      \State \textbf{Inputs}: Environment observation $e$, program id $i$, arguments $a$, stop threshold $\alpha$
      \Function{Run}{$i, a$}
        \State $h \gets \mathbf{0}$, $r \gets 0$, $p \gets M^{\texttt{prog}}_{i,:}$ \Comment{Init LSTM and return probability.}
        \While{$r < \alpha$} 
          \State $s \gets f_{enc}(e, a)$, $h \gets f_{lstm}(s, p, h)$ \Comment{Feed-forward NPI one step.}
          \State $r \gets f_{end}(h), k \gets f_{prog}(h), a_2 \gets f_{arg}(h)$ 
          \State $i_2 \gets \underset{j = 1..N}{\text{arg max}} (M^{\texttt{key}}_{j,:})^{T}k$    \Comment{Decide the next program to run.}
          \If{$i == $ ACT}
            $e \gets f_{env}(e,p,a)$ \Comment{Update the environment based on ACT.}
          \Else \textsc{ Run}($i_2, a_2$) \Comment{Run subprogram $i_2$ with arguments $a_2$}
          \EndIf
        \EndWhile
      \EndFunction
    \end{algorithmic}
  \end{algorithm}
%

Each task has a set of actions that affect the environment.
For example, in addition there are LEFT and RIGHT actions that move a specified pointer, and a WRITE action which writes a value at a specified location.
These actions are encapsulated into a general-purpose ACT program shared across tasks, and the concrete action to be taken is indicated by the NPI-generated arguments $a_{t}$.
%
%
%

Note that the core LSTM module of our NPI representation is completely agnostic to the data modality used to produce the state encoding.
As long as the same fixed-length embedding is extracted, the same module can in practice route between programs related to sorting arrays just as easily as between programs related to rotating 3D objects.
In the experimental sections, we provide details of the modality-specific deep neural networks that we use to produce these fixed-length state vectors.
%
%
%

%
%
%
%
%
%
\vspace{-0.1in}
\subsection{Training}
\vspace{-0.05in}
To train we use execution traces $\xi_t^{inp} : \{ e_t, i_t, a_t \}$ and $\xi_t^{out} : \{ i_{t+1}, a_{t+1}, r_t \}, t = 1, ... T$, where $T$ is the sequence length.
Program IDs $i_t$ and $i_{t+1}$ are row-indices in $M^{\texttt{key}}$ and $M^{\texttt{prog}}$ of the programs to run at time $t$ and $t+1$, respectively.
We propose to directly maximize the probability of the correct execution trace output $\xi^{out}$ conditioned on $\xi^{inp}$:
\begin{align}
\theta^{*} &= \underset{\theta}{\text{arg max}}\sum_{(\xi^{inp},\xi^{out})} \log P(\xi^{out} | \xi^{inp}; \theta)
\end{align}
where $\theta$ are the parameters of our model.
Since the traces are variable in length depending on the input, we apply the chain rule to model the joint probability over $\xi^{out}_{1} , ... , \xi^{out}_{T}$ as follows:
\begin{align}
\label{eq:prob}
\log P(\xi_{out} | \xi_{inp}; \theta) &= \sum_{t=1}^{T} \log P(\xi^{out}_{t} | \xi^{inp}_{1}, ..., \xi^{inp}_{t}; \theta)
\end{align}
Note that for many problems the input history $\xi^{inp}_{1}, ..., \xi^{inp}_{t}$ is critical to deciding future actions because the environment observation at the current time-step $e_t$ alone does not contain enough information.
The hidden unit activations of the LSTM in NPI are capable of capturing these temporal dependencies.
The single-step conditional probability in equation~(\ref{eq:prob}) can be factorized into three further conditional distributions, corresponding to predicting the next program, next arguments, and whether to halt execution:
\begin{align}
\log P(\xi^{out}_{t} | \xi^{inp}_{1}, ..., \xi^{inp}_{t}) =
\log P(i_{t+1} | h_t) + \log P(a_{t+1} | h _t) + \log P(r_t | h_t) 
\end{align}
where $h_t$ is the output of $f_{lstm}$ at time $t$, carrying information from previous time steps.
%
%
We train by gradient ascent on the likelihood in equation~(\ref{eq:prob}).

%
%
%
%
%
%
We used an adaptive curriculum in which training examples for each mini-batch are fetched with frequency proportional to the model's current prediction error for the corresponding program.
Specifically, we set the sampling frequency using a softmax over average prediction error across all programs, with configurable temperature.
Every 1000 steps of training we re-estimated these prediction errors.
Intuitively, this forces the model to focus on learning the program for which it currently performs worst in executing.
We found that the adaptive curriculum immediately worked much better than our best-performing hand-designed curriculum, allowing a multi-task NPI to achieve comparable performance to single-task NPI on all tasks.

We also note that our program has a distinct memory advantage over basic LSTMs because all subprograms can be trained in parallel.
For programs whose execution length grows \emph{e.g.} quadratically with the input sequence length, an LSTM will by highly constrained by device memory to train on short sequences.
By exploiting compositionality, an effective curriculum can often be developed with sublinear-length subprograms, enabling our NPI model to train on order of magnitude larger sequences than the LSTM.
\vspace{-0.1in}
\section{Experiments}
\label{sec:experiments}
\vspace{-0.1in}
%
%
This section describes the environment and state encoder function for each task, and shows example outputs and prediction accuracy results.
For all tasks, the core LSTM had two layers of size 256.
We trained the NPI using the ADAM solver~\citep{kingma2014adam} with base learning rate $0.0001$, batch size 1, and decayed the learning rate by a factor of $0.95$ every 10,000 steps.

\vspace{-0.05in}
\subsection{Task and environment descriptions}
\vspace{-0.05in}
In this section we provide an overview of the tasks used to evaluate our model. Table~\ref{tab:programs} in the appendix provides a full listing of all the programs and subprograms learned by our model. 
%
%
\subsubsection*{\textbf{Addition}}
%
The task in this environment is to read in the digits of two base-10 numbers and produce the digits of the answer.
Our goal is to teach the model the standard (at least in the US) grade school algorithm of adding, in which one works from right to left applying single-digit add and carry operations.

\begin{figure}[t!]
    \caption{Illustration of the addition environment used in our experiments.}
    \begin{center}
    \vspace{-0.1in}
    \begin{subfigure}{.35\textwidth}
        \center
        \includegraphics[width=0.72\linewidth]{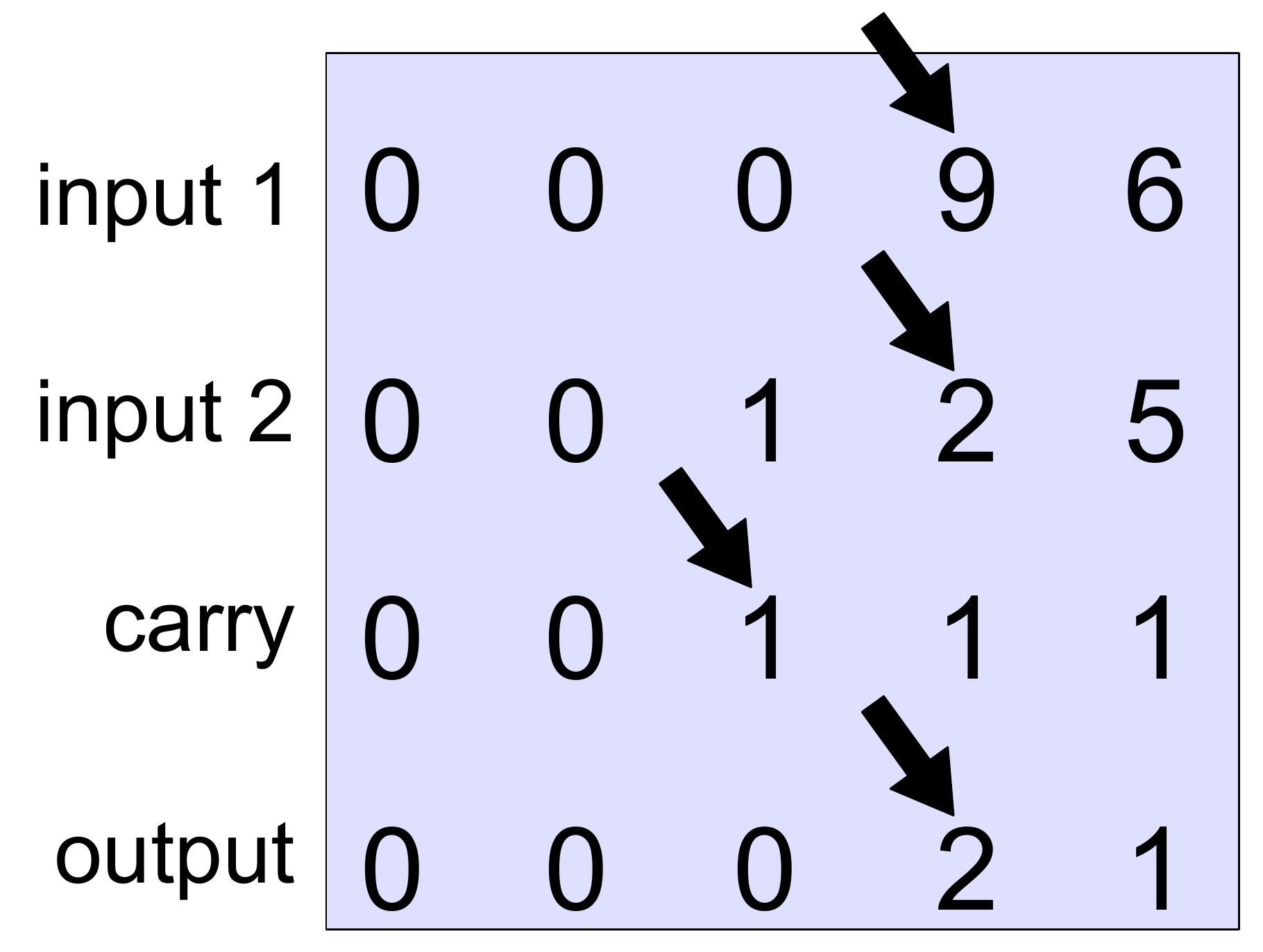}
        \caption{Example scratch pad and pointers used for computing ``96 + 125 = 221''. Carry step is being implemented.}
        \label{fig:addition}
    \end{subfigure}
    \hspace{0.2in}
    \begin{subfigure}{.55\textwidth}
        \center
        \includegraphics[width=0.82\linewidth]{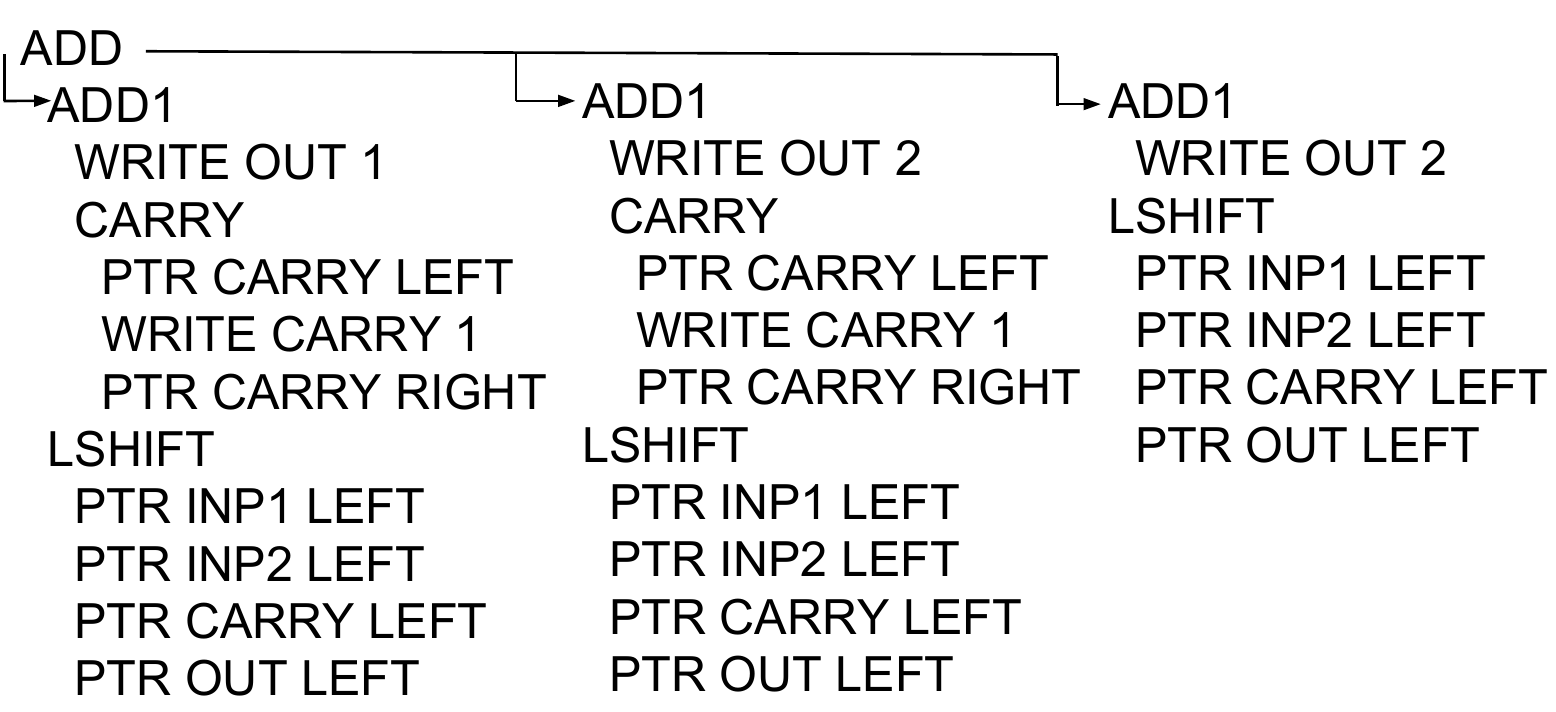}
        \caption{Actual trace of addition program generated by our model on the problem shown to the left. Note that we substituted the ACT calls in the trace with more human-readable steps.}
        \label{fig:addition_trace}
    \end{subfigure}
    \end{center}
    \vspace{-0.25in}
\end{figure}

In this environment, the network is endowed with a ``scratch pad'' with which to store intermediate computations; \emph{e.g.} to record carries.
There are four pointers; one for each of the two input numbers, one for the carry, and another to write the output.
At each time step, a pointer can be moved left or right, or it can record a value to the pad.
Figure~\ref{fig:addition} illustrates the environment of this model, and Figure~\ref{fig:addition_trace} provides a real execution trace generated by our model.

For the state encoder $f_{enc}$, the model is allowed a view of the scratch pad from the perspective of each of the four pointers.
That is, the model sees the current values at pointer locations of the two inputs, the carry row and the output row, as 1-of-K encodings, where K is 10 because we are working in base 10.
%
%
We also append the values of the input argument tuple $a_{t}$:
\begin{align}
f_{enc}(Q,i_1,i_2,i_3,i_4,a_{t}) =
MLP([ Q(1,i_1), Q(2,i_2), Q(3,i_3), Q(4,i_4),
a_{t}(1), a_{t}(2), a_{t}(3) ])
\end{align}
where $Q \in \mathbb{R}^{4 \times N \times K}$, and $i_1, ..., i_4$ are pointers, one per scratch pad row. The first dimension of $Q$ corresponds to scratch pad rows, $N$ is the number of columns (digits) and $K$ is the one-hot encoding dimension.
To begin the ADD program, we set the initial arguments to a default value and initialize all pointers to be at the rightmost column.
The only subprogram with non-default arguments is ACT, in which case the arguments indicate an action to be taken by a specified pointer.
%
\subsubsection*{\textbf{Sorting}}
\begin{figure}[t!]
    \caption{Illustration of the sorting environment used in our experiments.}
    \begin{center}
    \vspace{-0.1in}
    \begin{subfigure}{.35\textwidth}
        \center
        \includegraphics[width=0.7\linewidth]{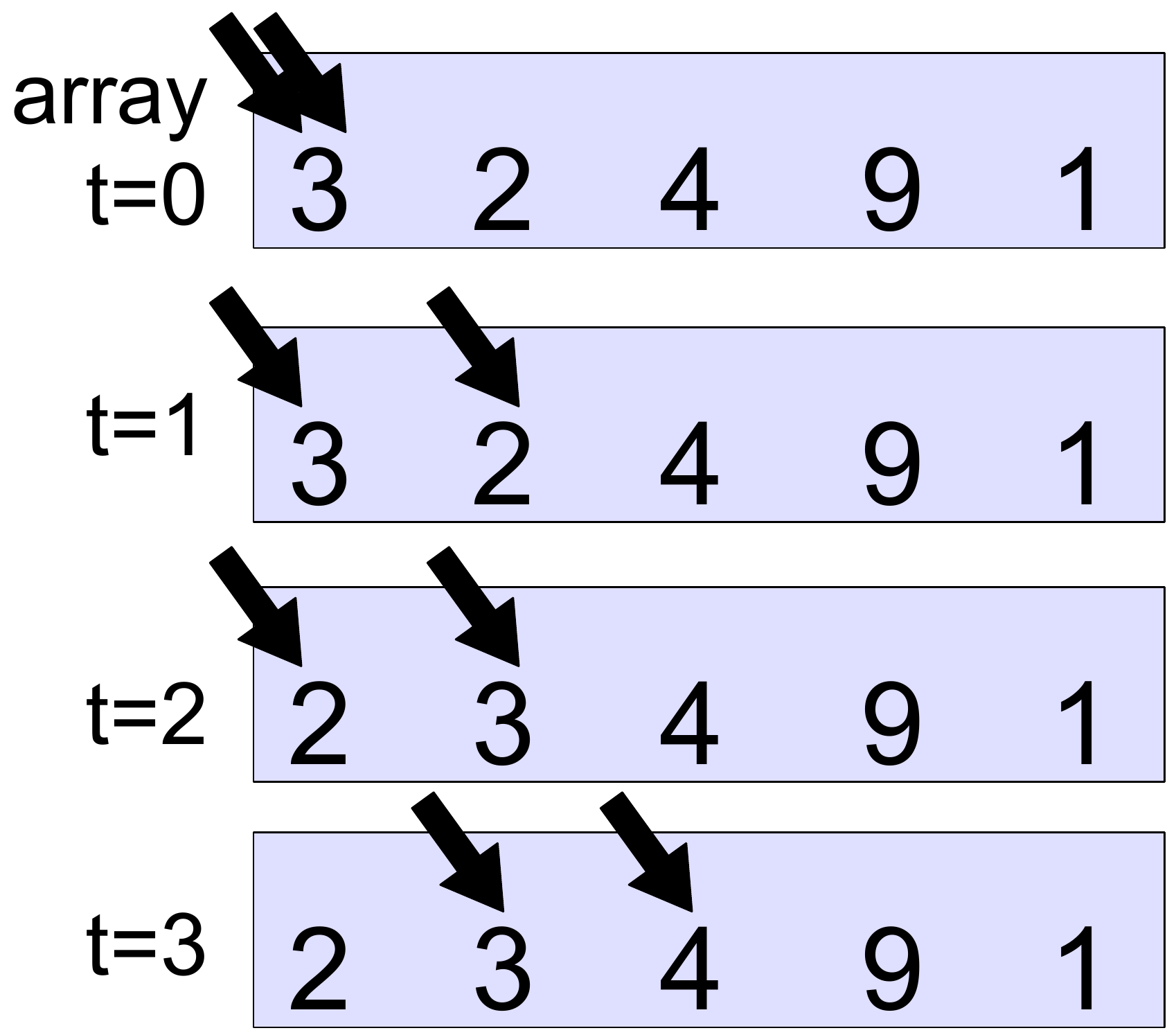}
        \caption{Example scratch pad and pointers used for sorting. Several steps of the BUBBLE subprogram are shown.}
        \label{fig:sorting}
    \end{subfigure}
    \hspace{0.1in}
    \begin{subfigure}{.55\textwidth}
        \center
        \includegraphics[width=0.73\linewidth]{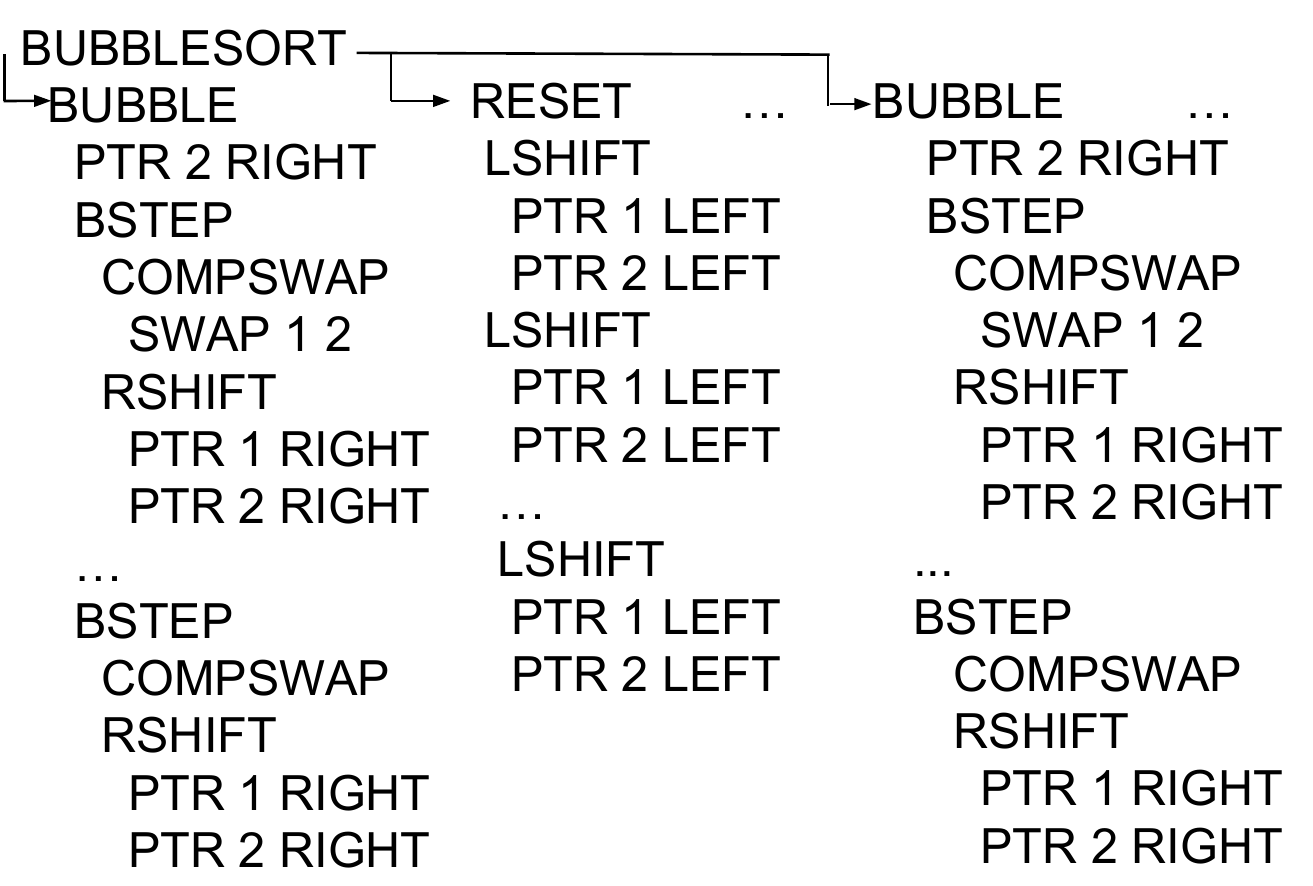}
        \caption{Excerpt from the trace of the learned bubblesort program.}
        \label{fig:sorting_trace}
    \end{subfigure}
    \vspace{-0.1in}
    \label{fig:bubblesort}
    \vspace{-0.1in}
    \end{center}
    \vspace{-0.1in}
\end{figure}

%
In this section we apply our model to a setting with potentially much longer execution traces: sorting an array of numbers using bubblesort.
As in the case of addition we can use a scratch pad to store intermediate states of the array.
We define the encoder as follows:
%
\begin{align}
f_{enc}(Q,i_1,i_2,a_{t}) = MLP([ Q(1,i_1), Q(1,i_2), a_{t}(1), a_{t}(2), a_{t}(3) ])
\end{align}
where $Q \in \mathbb{R}^{1 \times N \times K}$ is the pad, $N$ is the array length 
and K is the array entry embedding dimension.
Figure~\ref{fig:bubblesort} shows an example series of array states and an excerpt of an execution trace.

\subsubsection*{\textbf{Canonicalizing 3D models}}
%
%
We also apply our model to a vision task with a very different perceptual environment - pixels.
Given a rendering of a 3D car, we would like to learn a visual program that ``canonicalizes'' the model with respect to its pose.
Whatever the starting position, the program should generate a trajectory of actions that delivers the camera to the target view, \emph{e.g.} frontal pose at a $15^{\circ}$ elevation.
For training data, we used renderings of the 3D car CAD models from~\citep{FidlerNIPS12}.

This is a nontrivial problem because different starting positions will require quite different trajectories to reach the target.
Further complicating the problem is the fact that the model will need to generalize to different car models than it saw during training.

We again use a scratch pad, but here it is a very simple read-only pad that only contains a target camera elevation and azimuth -- \emph{i.e.}, the ``canonical pose''.
Since observations come in the form of image pixels, we use a convolutional neural network $f_{CNN}$ as the image encoder:
%
\begin{align}
f_{enc}(Q,x,i_1,i_2,a_{t}) = MLP([ Q(1,i_1), Q(2,i_2), f_{CNN}(x), a_{t}(1), a_{t}(2), a_{t}(3) ])
\end{align}
where $x \in \mathbb{R}^{H \times W \times 3}$ is a car rendering at the current pose, $Q \in \mathbb{R}^{2 \times 1 \times K}$ is the pad containing canonical azimuth and elevation, $i_1, i_2$ are the (fixed at 1) pointer locations, and $K$ is the one-hot encoding dimension of pose coordinates.
We set $K = 24$ corresponding to $15^{\circ}$ pose increments.

Note, critically, that our NPI model only has access to pixels of the rendering and the target pose, and is not provided the pose of query frames.
We are also aware that one solution to this problem would be to train a pose classifier network and then find the shortest path to canonical pose via classical methods.
That is also a sensible approach.
However, our purpose here is to show that our method generalizes beyond the scratch pad domain to detailed images of 3D objects, and also to other environments with a single multi-task model.
\vspace{-0.05in}
\subsection{Sample complexity and generalization}
\vspace{-0.05in}
Both LSTMs and Neural Turing Machines can learn to perform sorting to a limited degree, although they have not been shown to generalize well to much longer arrays than were seen during training.
However, we are interested not only in whether sorting can be accomplished, but whether a particular sorting algorithm (e.g. bubblesort) can be learned by the model, and how effectively in terms of sample complexity and generalization.

We compare the generalization ability of our model to a flat sequence-to-sequence LSTM~\citep{sutskever2014sequence}, using the same number of layers (2) and hidden units (256).
Note that a flat~\footnote{By flat in this case, we mean non-compositional, not making use of subprograms, and only making calls to ACT in order to swap values and move pointers.} version of NPI could also learn sorting of short arrays, but because bubblesort runs in $O(N^2)$ for arrays of length $N$, the execution traces quickly become far too long to store the required number of LSTM states in memory. 
Our NPI architecture can train on much larger arrays by exploiting compositional structure; the memory requirements of any given subprogram can be restricted to $O(N)$. 
%
%

\begin{figure}[h!]
  \begin{minipage}[c]{0.49\textwidth}
    \includegraphics[width=\textwidth]{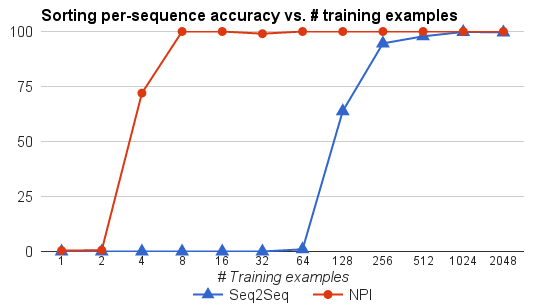}
    \vspace{-0.2in}
    \caption{\textbf{Sample complexity}. Test accuracy of sequence-to-sequence LSTM versus NPI on length-20 arrays of single-digit numbers. Note that NPI is able to mine and train on subprogram traces from each bubblesort example.}
    \label{fig:sc_fat}
  \end{minipage}\hfill
  \begin{minipage}[c]{0.49\textwidth}
    \includegraphics[width=\textwidth]{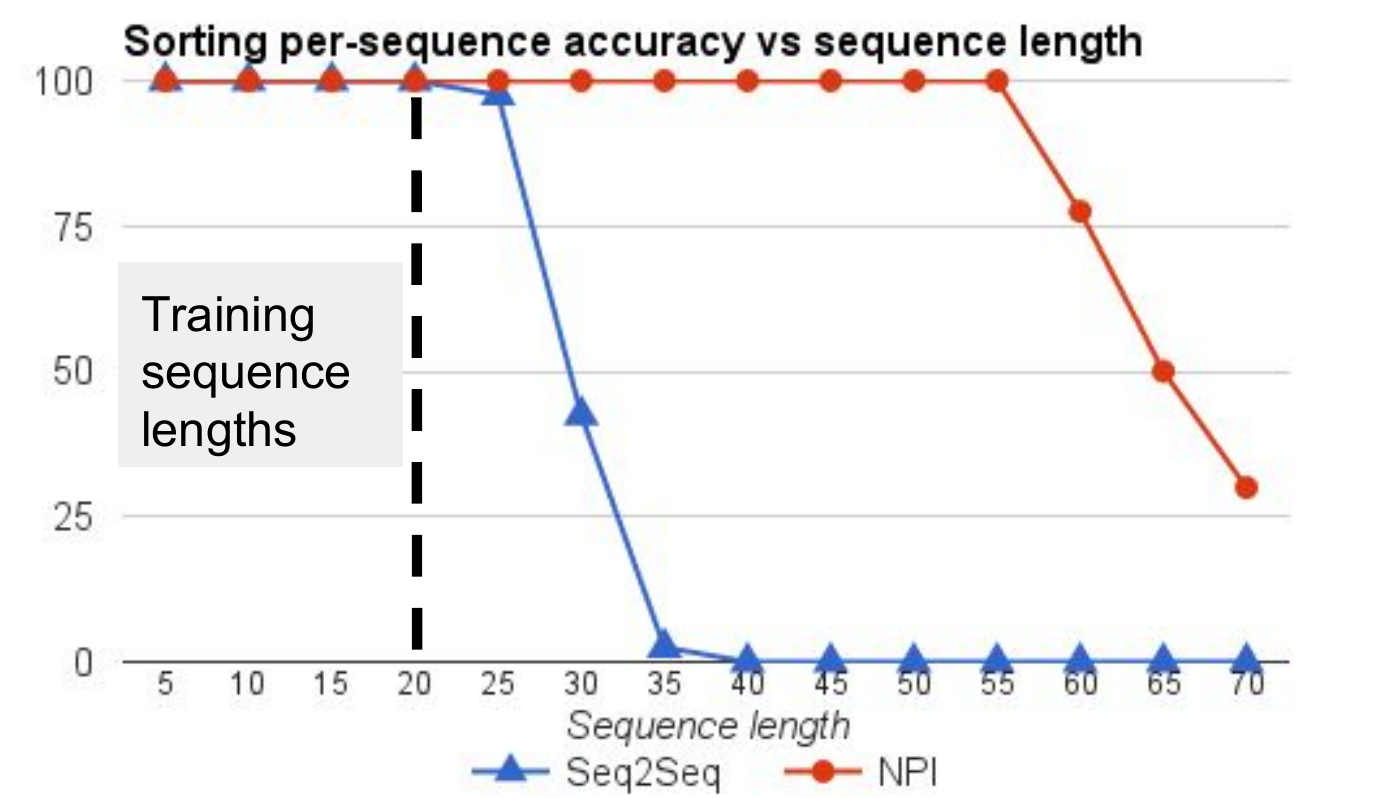}
    \vspace{-0.2in}
    \caption{\textbf{Strong vs. weak generalization}. Test accuracy of sequence-to-sequence LSTM versus NPI on varying-length arrays of single-digit numbers. Both models were trained on arrays of single-digit numbers up to length 20.}
\label{fig:sc_gen}
  \end{minipage}
  
  \vspace{-0.16in}
\end{figure}

%
%
%

%
%
%
%

%
A strong indicator of whether a neural network has learned a program well is whether it can run the program on inputs of previously-unseen sizes.
To evaluate this property, we train both the sequence-to-sequence LSTM and NPI to perform bubblesort on arrays of single-digit numbers from length 2 to length 20.
%
%
Compared to fixed-length inputs this raises the challenge level during training, but in exchange we can get a more flexible and generalizable sorting program.

To handle variable-sized inputs, the state representation must have some information about input sequence length and the number of steps taken so far.
For example, the main BUBBLESORT program naturally needs to call its helper function BUBBLE a number of times dependent on the sequence length.
We enable this in our model by adding a third pointer that acts as a counter; each time BUBBLE is called the pointer is advanced by one step.
The scratch pad environment also provides a bit indicating whether a pointer is at the start or end of a sequence, equivalent in purpose to end tokens used in a sequence-to-sequence model.
%
%
%
%

%
%
For each length, we provided 64 example bubblesort traces, for a total of 1,216 examples.
Then, we evaluated whether the network can learn to sort arrays beyond length 20.
We found that the trained model generalizes well, and is capable of sorting arrays up to size 60; see Figure~\ref{fig:sc_gen}.
At 60 and beyond, we observed a failure mode in which sweeps of pointers across the array would take the wrong number of steps, suggesting that the limiting performance factor is related to counting.
%
In stark contrast, when provided with the 1,216 examples, the sequence-to-sequence LSTMs fail to generalize beyond arrays of length 25 as shown in Figure~\ref{fig:sc_gen}.

To study sample complexity further, we fix the length of the arrays to 20 and vary the number of training examples. We see in Figure~\ref{fig:sc_fat} that NPI starts learning with 2 examples and is able to sort almost perfectly with only 8 examples. The sequence-to-sequence model on the other hand requires 64 examples to start learning and only manages to sort well with over 250 examples.  

\begin{figure}[h!]
\vspace{0.1in}
\center
\includegraphics[width=\linewidth]{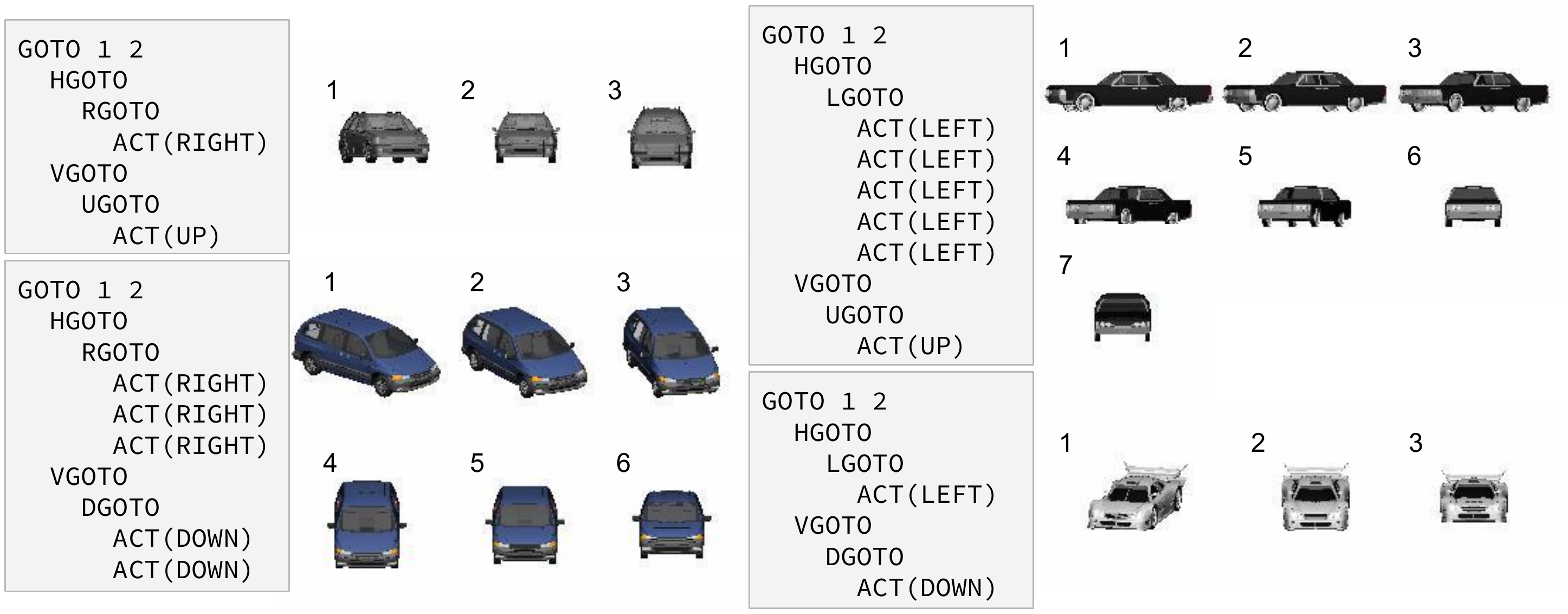}
\vspace{-0.25in}
\caption{Example canonicalization of several different test set cars. The network is able to generate and execute the appropriate plan based on the starting car image. This NPI was trained on trajectories starting at azimuth $(-75^{\circ} ... 75^{\circ})$ , elevation $(0^{\circ} ... 60^{\circ})$ in $15^{\circ}$ increments. The training trajectories target azimuth $0^{\circ}$ and elevation $15^{\circ}$, as in the generated traces above.}
\label{fig:canonicalize}
\vspace{-0.1in}
\end{figure}

Figure~\ref{fig:canonicalize} shows several example canonicalization trajectories generated by our model, starting from the leftmost car.
The image encoder was a convolutional network with three passes of stride-2 convolution and pooling, trained on renderings of size $128 \times 128$.
The canonical target pose in this case is frontal with $15^{\circ}$ elevation. At test time, 
from an initial rendering, NPI is able to canonicalize cars of varying appearance from multiple starting positions. Importantly, it can generalize to car appearances not encountered in the training set as shown in Figure~\ref{fig:canonicalize}.   
\vspace{-0.05in}
\subsection{Learning new programs with a fixed core}
\vspace{-0.05in}
\label{sec:fixed}
One challenge for continual learning of neural-network-based agents is that training on new tasks and experiences can lead to degraded performance in old tasks.
The learning of new tasks may require that the network weights change substantially, so care must be taken to avoid catastrophic forgetting~\citep{McCloskey:1989,OReilly:2014}.
Using NPI, one solution is to fix the weights of the core routing module, and only make sparse updates to the program memory.

When adding a new program the core module's routing computation will be completely unaffected; all the learning for a new task occurs in program embedding space.
Of course, the addition of new programs to the memory adds a new choice of program at each time step, and an old program could mistakenly call a newly added program.
To overcome this, when learning a new set of program vectors with a fixed core, in practice we train not only on example traces of the new program, but also traces of existing programs.
Alternatively, a simpler approach is to prevent existing programs from calling subsequently added programs, allowing addition of new programs without ever looking back at training data for known programs.
In either case, note that \emph{only the memory slots of the new programs} are updated, and all other weights, including other program embeddings, are fixed.

Table~\ref{tab:multitask} shows the result of adding a maximum-finding program MAX to a multitask NPI trained on addition, sorting and canonicalization.
MAX first calls BUBBLESORT and then a new program RJMP, which moves pointers to the right of the sorted array, where the max element can be read.
During training we froze all weights except for the two newly-added program embeddings.
We find that NPI learns MAX perfectly without forgetting the other tasks. In particular, after training a single multi-task model as outlined in the following section, learning the MAX program with this fixed-core multi-task NPI results in no performance deterioration for all three tasks.  
\vspace{-0.1in}
\subsection{Solving multiple tasks with a single network}
\vspace{-0.05in}
\label{sec:multiple}
In this section we perform a controlled experiment to compare the performance of a multi-task NPI with several single-task NPI models.
Table~\ref{tab:multitask} shows the results for addition, sorting and canonicalizing 3D car models.
We trained and evaluated on 10-digit numbers for addition, length-5 arrays for sorting, and up to four-step trajectories for canonicalization.
As shown in Table~\ref{tab:multitask}, one multi-task NPI can learn all three programs (and necessarily the 21 subprograms) with comparable accuracy compared to each single-task NPI.
\vspace{2mm}
\begin{table}[h!]
  \vspace{-0.2in}
  \begin{minipage}[c]{0.5\textwidth}
    \vspace{-0.1in}
    \begin{tabular}{l|c|c|c}
    \hline
    \textbf{Task} & \textbf{Single} & \textbf{Multi} & \textbf{+ Max} \\ \hline
    \hline
    Addition & 100.0 & 97.0 &  97.0 \\ \hline
    Sorting & 100.0 & 100.0 & 100.0 \\ \hline
    Canon. seen car & 89.5 & 91.4 & 91.4\\ \hline
    Canon. unseen & 88.7 & 89.9 & 89.9\\ \hline
    Maximum & - & - & 100.0 \\ \hline
    \end{tabular}
  \end{minipage}\hfill
  \begin{minipage}[c]{0.48\textwidth}
    \vspace{0.1in}
    \caption{Per-sequence \% accuracy. ``+ Max'' indicates performance after addition of the additional max-finding subprograms to memory. ``unseen'' uses a test set with disjoint car models from the training set, while ``seen car'' uses the same car models but different trajectories.}
       \label{tab:multitask}
  \end{minipage}
  \vspace{-0.3in}
\end{table}
\vspace{-0.1in}
\section{Conclusion}
\vspace{-0.1in}
We have shown that the NPI can learn programs in very dissimilar environments with different affordances. In the context of sorting we showed that NPI exhibits very strong generalization in comparison to sequence-to-sequence LSTMs. We also showed how a trained NPI with a fixed core can continue to learn new programs without forgetting already learned programs.
%
\subsubsection*{Acknowledgments}
We sincerely thank Arun Nair and Ed Grefenstette for helpful suggestions.
\bibliography{references}
\bibliographystyle{iclr2016_conference}
\clearpage
\section{Appendix}
\subsection{Listing of learned programs}
Below we list the programs learned by our model:
\begin{table}[h!]
  \begin{center}
  {\small 
  \begin{tabular}{ | l | l | l |}
    \hline
    \textbf{Program} & \textbf{Descriptions} & \textbf{Calls} \\ \hline
    \textsc{ADD} & Perform multi-digit addition & \textsc{ADD1}, \textsc{LSHIFT} \\ \hline
    \textsc{ADD1} & Perform single-digit addition & \textsc{ACT}, \textsc{CARRY} \\ \hline
    \textsc{CARRY} & Mark a 1 in the carry row one unit left & \textsc{ACT} \\ \hline
    \textsc{LSHIFT} & Shift a specified pointer one step left & \textsc{ACT} \\ \hline
    \textsc{RSHIFT} & Shift a specified pointer one step right & \textsc{ACT} \\ \hline
    \textsc{ACT} & Move a pointer or write to the scratch pad & - \\
    \hline
    \hline
    \textsc{BUBBLESORT} & Perform bubble sort (ascending order) & \textsc{BUBBLE}, \textsc{RESET} \\ \hline
    \textsc{BUBBLE} & Perform one sweep of pointers left to right & \textsc{ACT}, \textsc{BSTEP} \\ \hline
    \textsc{RESET} & Move both pointers all the way left & \textsc{LSHIFT} \\ \hline
    \textsc{BSTEP} & Conditionally swap and advance pointers & \textsc{COMPSWAP}, \textsc{RSHIFT} \\ \hline
    \textsc{COMPSWAP} & Conditionally swap two elements & \textsc{ACT} \\ \hline
    \textsc{LSHIFT} & Shift a specified pointer one step left & \textsc{ACT} \\ \hline
    \textsc{RSHIFT} & Shift a specified pointer one step right & \textsc{ACT} \\ \hline
    \textsc{ACT} & Swap two values at pointer locations or move a pointer & - \\
    \hline
    \hline
    \textsc{GOTO} & Change 3D car pose to match the target & \textsc{HGOTO}, \textsc{VGOTO} \\ \hline
    \textsc{HGOTO} & Move horizontally to the target angle & \textsc{LGOTO}, \textsc{RGOTO} \\ \hline
    \textsc{LGOTO} & Move left to match the target angle & \textsc{ACT} \\ \hline
    \textsc{RGOTO} & Move right to match the target angle & \textsc{ACT} \\ \hline
    \textsc{VGOTO} & Move vertically to the target elevation & \textsc{UGOTO}, \textsc{DGOTO} \\ \hline
    \textsc{UGOTO} & Move up to match the target elevation & \textsc{ACT} \\ \hline
    \textsc{DGOTO} & Move down to match the target elevation & \textsc{ACT} \\ \hline
    \textsc{ACT} & Move camera $15^{\circ}$ up, down, left or right & - \\
    \hline
    \hline
    \textsc{RJMP} & Move all pointers to the rightmost posiiton & \textsc{RSHIFT} \\ \hline
    \textsc{MAX} & Find maximum element of an array & \textsc{BUBBLESORT},\textsc{RJMP} \\
    \hline
  \end{tabular}
  }
  \end{center}
  \vspace{-0.1in}
  \caption{Programs learned for addition, sorting and 3D car canonicalization. Note the the \textsc{ACT} program has a different effect depending on the environment and on the passed-in arguments.}
  \label{tab:programs}
\vspace{-0.1in}
\end{table}
\vspace{-0.1in}
\subsection{Generated execution trace of BUBBLESORT}
\begin{figure}[h!]
    \caption{Generated execution trace from our trained NPI sorting the array [9,2,5].}
    \vspace{-0.2in}
    \center
    \includegraphics[width=0.6\linewidth]{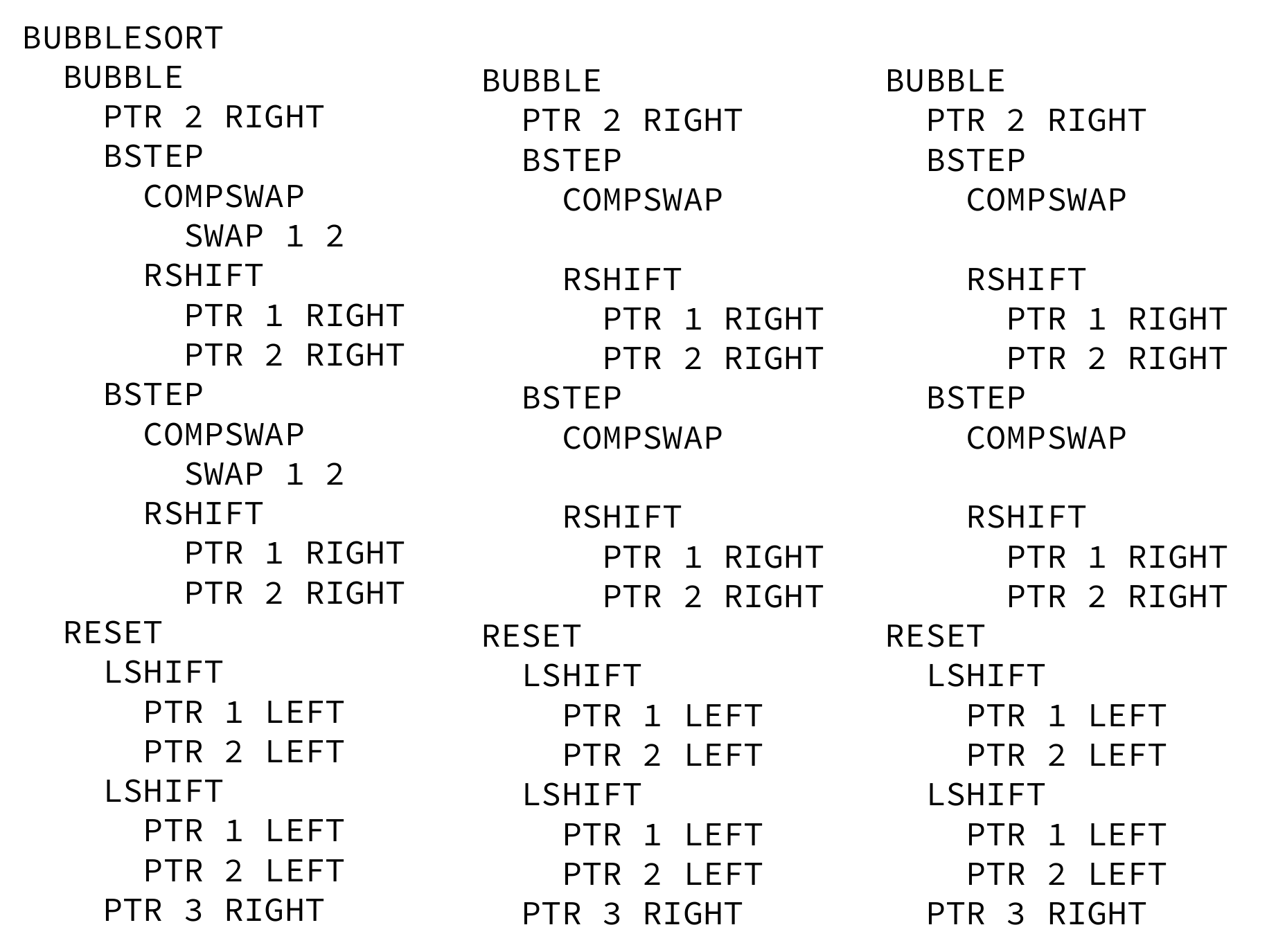}
    \label{fig:bubblesort_trace}
    \vspace{-0.2in}
\end{figure}

\vspace{-0.1in}
%
%
Figure~\ref{fig:bubblesort_trace} shows the sequence of program calls for BUBBLESORT. 
Pointers 1 and 2 are used to implement the ``bubble'' operation involving the comparison and swapping of adjacent array elements.
The third pointer (referred to in the trace as ``PTR 3'') is used to count the number of calls to BUBBLE.
After every call to RESET the swapping pointers are moved to the beginning of the array and the counting pointer is advanced by 1.
When it has reached the end of the scratch pad, the model learns to halt execution of BUBBLESORT.

\subsection{Additional experiment on addition generalization}
Based on reviewer feedback, we conducted an additional comparison of NPI and sequence-to-sequence models for the addition task, to evaluate the generalization ability.
we implemented addition in a sequence to sequence model, training to model sequences of the following form, e.g. for ``90 + 160 = 250'' we represent the sequence as:

\begin{verbatim}
90X160X250
\end{verbatim}

For the simple Seq2Seq baseline above (same number of LSTM layers and hidden units as NPI), we observed that the model could predict one or two digits reliably, but did not generalize even up to 20-digit addition.
However, we are aware that others have gotten multi-digit addition of the above form to work to some extent with curriculum learning~\citep{zaremba2014learning}.
In order to make a more competitive baseline, we helped Seq2Seq in two ways: 1) reverse input digits and stack the two numbers on top of each other to form a 2-channel sequence, and 2) reverse input digits and generate reversed output digits immediately at each time step.

In the approach of 1), the seq2seq model schematically looks like this:
\begin{verbatim}
output:  XXXX250
input 1: 090XXXX
input 2: 061XXXX
\end{verbatim}
In the approach of 2), the sequence looks like this:
\begin{verbatim}
output:  052
input 1: 090
input 2: 061
\end{verbatim}

Both 1) which we call “s2s-stacked” and 2) which we call “s2s-easy” are much stronger competitors to NPI than even the proposed addition baseline.
We compare the generalization performance of NPI to these baselines in the figure below:

\begin{figure}[h!]
   \begin{center}
   \includegraphics[width=0.8\linewidth]{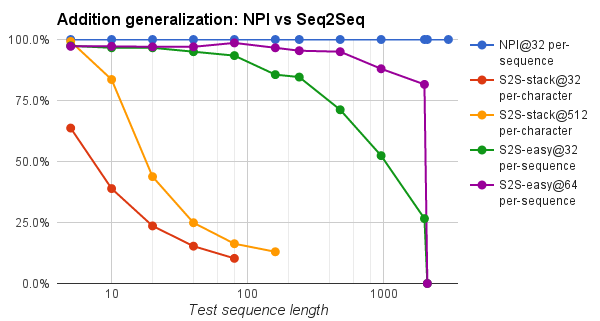}
   \caption{Comparing NPI and Seq2Seq variants on addition generalization to longer sequences.}
   \vspace{-0.1in}
   \end{center}
   \label{fig:addition_rebuttal}
\end{figure}

We found that NPI trained on 32 examples for problem lengths 1,...,20 generalizes with 100\% accuracy to all the lengths we tried (up to 3000).
s2s-easy trained on twice as many examples generalizes to just over length 2000 problems.
s2s-stacked barely generalizes beyond 5, even with far more data.
This suggests that locality of computation makes a large impact on generalization performance.
Even when we carefully ordered and stacked the input numbers for Seq2Seq, NPI still had an edge in performance.
In contrast to Seq2Seq, NPI is taught (supervised for now) to move its pointers so that the key operations (e.g. single digit add, carry) can be done using only local information, and this appears to help generalization.

\end{document}